\documentclass[runningheads]{llncs}
\usepackage{color,xcolor}
\usepackage{multirow}
\usepackage{booktabs}
\usepackage{amsmath}
\usepackage{amssymb}
\usepackage{url}
\usepackage{algpseudocode}
\usepackage{amsfonts}
\usepackage[ruled,linesnumbered]{algorithm2e}
\usepackage{multicol}     
\usepackage{multirow}
\usepackage[misc]{ifsym}
\usepackage{color}
\usepackage[T1]{fontenc}
\usepackage{cite}
\usepackage{subfigure}

\usepackage{graphicx}
\begin{document}
\title{SubGraph Networks based Entity Alignment for Cross-lingual Knowledge Graph}
\titlerunning{SubGraph Networks based Entity Alignment for Cross-lingual KG}
%
\author{Shanqing Yu\inst{1,2}\textsuperscript{(\Letter)} \and
        Shihan Zhang\inst{1,2} \and
        Jianlin Zhang\inst{1,2} \and
        Jiajun Zhou\inst{1,2}  \and \\
        Qi Xuan\inst{1,2}  \and
        Bing Li\inst{3} \and
        Xiaojuan Hu\inst{4}}

\authorrunning{S. Yu et al.}
%
\institute{
    Institute of Cyberspace Security, Zhejiang University of Technology, \\Hangzhou 310023, China \\ \and
    College of Information Engineering, Zhejiang University of Technology, \\Hangzhou 310023, China\\ \and
    Hangzhou Zhongao Technology Company Limited, \\Hangzhou 310000, China\\ \and
    Zhejiang Economic Information Center, \\Hangzhou 310000,  China\\
    \email{yushanqing@zjut.edu.cn}   
}
%
\maketitle              
\begin{abstract}
Entity alignment is the task of finding entities representing the same real-world object in two knowledge graphs(KGs). Cross-lingual knowledge graph entity alignment aims to discover the cross-lingual links in the multi-language KGs, which is of great significance to the NLP applications and multi-language KGs fusion. In the task of aligning cross-language knowledge graphs, the structures of the two graphs are very similar, and the equivalent entities often have the same subgraph structure characteristics. The traditional GCN method neglects to obtain structural features through representative parts of the original graph  and the use of adjacency matrix is not enough to effectively represent the structural features of the graph. In this paper, we introduce the subgraph network (SGN) method into the GCN-based cross-lingual KG entity alignment method. In the method, we extracted the first-order subgraphs of the KGs to expand the structural features of the original graph to enhance the representation ability of the entity embedding and improve the alignment accuracy. Experiments show that the proposed method outperforms the state-of-the-art GCN-based method.

\end{abstract}
\section{Introduction}
The knowledge graph, which aims to organize human knowledge in a structured form, is playing an increasingly important role as an infrastructure in the field of artificial intelligence and natural language processing~\cite{aditya2018explicit}. A knowledge graph is a collection of a series of knowledge facts, usually represented by a triple of (head entity, relationship, tail entity). With the vigorous development of multilingual knowledge graphs such as YAGO~\cite{suchanek2008yago,rebele2016yago} , DBpedia~\cite{lehmann2015dbpedia} and BabelNet~\cite{navigli2012babelnet}, the structured knowledge provided by these KGs is widely used as a priori for applications such as knowledge base question answering~\cite{zhang2017variational} and language modeling~\cite{ahn2016neural}. However, these multilingual knowledge graphs are difficult to integrate effectively due to different language codes. In order to effectively utilize the knowledge contained in KGs from different languages, more and more researches are focused on the alignment of cross-language knowledge graphs which aim to automatically discover and match the equivalent entities in multilingual knowledge graphs.

The multilingual knowledge graphs contains rich cross-language links. These links match the equivalent entities in different languages KG, and play an important role in completing the multilingual graph. Traditional cross-language graph alignment methods usually use machine translation or define independent features of various languages to discover cross-language links. Such methods are difficult to apply on a large scale. In recent years, with the popularity of knowledge graph embedding methods, many alignment methods based on embedding have been proposed.

The knowledge graph embedding can map the entities and relationships in the graph to a continuous low-dimensional vector space, and retain some semantic information. In recent years, many representative works have appeared in the field of knowledge graph embedding. Among them, TransE~\cite{bordes2013translating} is one of the most classic knowledge graph embedding methods. It regards the relationship in the triple $(h, r, t)$ as the translation from the head entity vector to the tail entity vector, and makes each triple must satisfy $h + r \approx t$. The model learns embedding by minimizing the Euclidean distance between the vectors $h + r$ and $t$. The TransE model is simple and powerful, and excels in the tasks of link prediction and triple classification. However, due to the problem of insufficient expression ability of the model, it cannot handle the one-to-one, one-to-many, and many-to-many relationships well. Researchers have proposed some improved models based on TransE. For example, TransH~\cite{wang2014knowledge} solves the problem that the TransE model cannot handle one-to-many and many-to-one issues by projecting the vector representations of the head entity $h$ and tail entity $t$ corresponding to each relationship onto different hyperplanes. TransR~\cite{lin2015learning} adds entity attribute information during the embedding process. TransD~\cite{ji2015knowledge} solves the vector representation problem of multiple semantic relations after embedding through dynamic matrix. TransA~\cite{xiao2015transa} changes the distance metric in the loss function to Mahalanobis distance, and sets different weights for each dimension of learning. HyTE~\cite{dasgupta2018hyte} considers the time validity of the establishment of triples. The above models have improved the expressive ability of TransE to a certain extent.

JE~\cite{hao2016joint} uses an embedding-based method to map the entities into a low-dimensional space by given two knowledge graphs and a set of pre-aligned entities, and then matches the equivalent entities according to their embedding vectors. MTransE~\cite{chen2016multilingual} increases the embedding of relations, so training the model needs to provide pre-aligned triples in two knowledge graphs. JAPE~\cite{sun2017cross} adds entity attribute embedding on the basis of MTransE to enhance the alignment effect. Wang~\cite{wang2018cross} and others first transferred the method of graph convolutional neural network to knowledge graph for entity alignment. However, none of the existing models make full use of the subgraph network~\cite{xuan2019subgraph} information of the knowledge graph. Since the network structure of cross-linguistic knowledge graphs is very similar, the use of adjacency matrix is not enough to effectively represent the structural features of the graph. By extracting its subgraph features, it can effectively make up for its lack of features.

Based on the above situation, this paper proposes a cross-language knowledge graph alignment model combining subgraph features. This method combines the knowledge model and the alignment model to learn the representation of the multilingual knowledge graph. Among them, the knowledge model uses the graph convolutional neural network to encode the feature information of the entity to obtain the node-level embedding vector. The alignment model comprehensively considers the structural embedding, the subgraph structural embedding and attribute embedding of the KG to find the equivalent entity. The main contributions of this paper are as follows:
\begin{itemize}
\item [$\bullet$]A method is proposed to expand the structural features of the original KG by using the sub-graph features, and make the structural embedding vector more suitable for the discovery of equivalent entities.
\item [$\bullet$]Combining the SGN and the GCN method proves the effectiveness of the sub-graph feature embedding in the task of aligning entities of the structurally similar knowledge graph.
\end{itemize}

The remainder of this paper is organized as follows. In Section~\ref{sec:RelatedWork}, we present some basic methods for GCN, SGN and knowledge graph alignment based on embedding. In Section~\ref{sec:Methodology}, we introduce a method of entity alignment in cross-language knowledge graph based on subgraph network. In Section~\ref{sec: Experiments}, we introduce the dataset and experiment extensively. In Section~\ref{sec:Conclusion}, we conclude the paper.

\section{Related Work}
\label{sec:RelatedWork}
\subsection{GCN}
The convolutional neural network (GCN)~\cite{defferrard2016convolutional,kipf2016semi} is a neural network that can run directly on the graph data. In the method of this article, input the adjacency matrix of the graph and the feature vector of the entity to the GCN, and then the domain information of the entity can be encoded as a real-valued vector. In the task of aligning entities of the cross-language knowledge graph, equivalent entities generally have similar structural features. Therefore, GCN model can effectively aggregate these two types of information and map entities to a low-dimensional vector space. And make the equivalent entities have similar vector representations.
The GCN model consists of several fixed GCN layers, and its $l + 1$ layer depends on the output of the previous layer, with the specific convolution calculated as follows: 
\begin{equation}
\begin{split}
H^{(l+1)}=\sigma(\widehat{Q}^{-\frac{1}{2}}&\widehat{P}\widehat{Q}^{-\frac{1}{2}}H(l)W(l)),
\end{split}
\end{equation}
where $P$ is the adjacency matrix of $n x n$ and $n$ is the number of nodes. $\widehat{P}=P+I$, $I$ is the identity matrix, $\widehat{Q}$ is $\widehat{P}$ diagonal nodal matrix, $H{(l)}$ is the vertex feature matrix input to the first layer of the GCN model, $W(l)$ is the weight matrices of the neural network of the first layer of various features, and $\sigma$ is similar to RELU's nonlinear activation function.

\subsection{SGN}
Many real-world systems can be naturally represented by networks, such as collaborative networks~\cite{nguyen2018learning, xuan2017social} and social networks~\cite{kim2018social, fu2018link}. Knowledge graph also contains rich network structure. The entities in KG can be regarded as nodes in the network, relationships can be regarded as edges in the network. Therefore, the knowledge graph can be regarded as a heterogeneous network in essence. 

Subgraphs are the basic components of a network, so studying the substructur of a network is an effective way to analyze a network~\cite{ullmann1976algorithm}. Recently, Word2vec~\cite{mikolov2013efficient}, DeepWalk~\cite{perozzi2014deepwalk} and other graph embedding algorithms have been widely used in node classification and other tasks. However, the embedded vector obtained by such models contains only local structure information around nodes, while ignoring the global structure information of the entire network. The SGN proposal effectively solves this problem, Xuan et al.~\cite{xuan2019subgraph} designed the algorithm used to build first-order and second-order SGN, and can be easily extended to build high-order subgraph network. Experiments show that the structural characteristics of SGN can complement the structural characteristics of the original network, so as to better carry out downstream tasks such as node classification and network classification.

\subsection{Knowledge Graph Alignment Based On Embedding}
Currently, the alignment of knowledge graph based on embedding can be divided into two categories: alignment based on the TransE model and alignment based on the GCN~\cite{kipf2016semi} model. This section will briefly introduce some of the most representative models and discuss the differences between them.

The JE~\cite{hao2016joint} model merges the pre-aligned entities into the same KG, and jointly learns the representation of multiple KGs in a unified vector space, and then aligns the entities in the knowledge map according to the embedded vector.

MTransE~\cite{chen2016multilingual} combines the knowledge model and the alignment model to align the knowledge graph. This method uses TransE to train two knowledge graphs separately, and encodes entities and relationships in independent spaces. Then use some pre-aligned triples for training, and the alignment model considers three cross-language alignment representations based on distance-based axis calibration, translation vector and linear transformation. The loss function of MTransE is the weighted sum of the knowledge model and the alignment model.

Sometimes, the structural information of the KG cannot be used to obtain a good alignment effect, so some scholars have proposed an alignment method combining attribute information. The JAPE~\cite{sun2017cross} model combines structure embedding and attribute embedding to match entities in different KGs. The model uses TransE to learn the structural features of the two KGs, and the Skip-gram model to capture attribute features, which improves the performance of MTransE to a certain extent.

The above methods all rely on the TransE model to learn entity embedding. Such methods are often limited by the expressive ability of the TransE model. Therefore, Wang~\cite{wang2018cross} and others proposed a method of graph convolutional neural network. This method uses the GCN model to embed the entities in different KGs into a unified vector space, so that the aligned entities are as close as possible. Unlike TransE-based models such as MTransE and JAPE, this method does not require pre-aligned triples, but only needs to focus on pre-aligned entities.

\begin{figure}[ht]
\centering
\includegraphics[width=\textwidth]{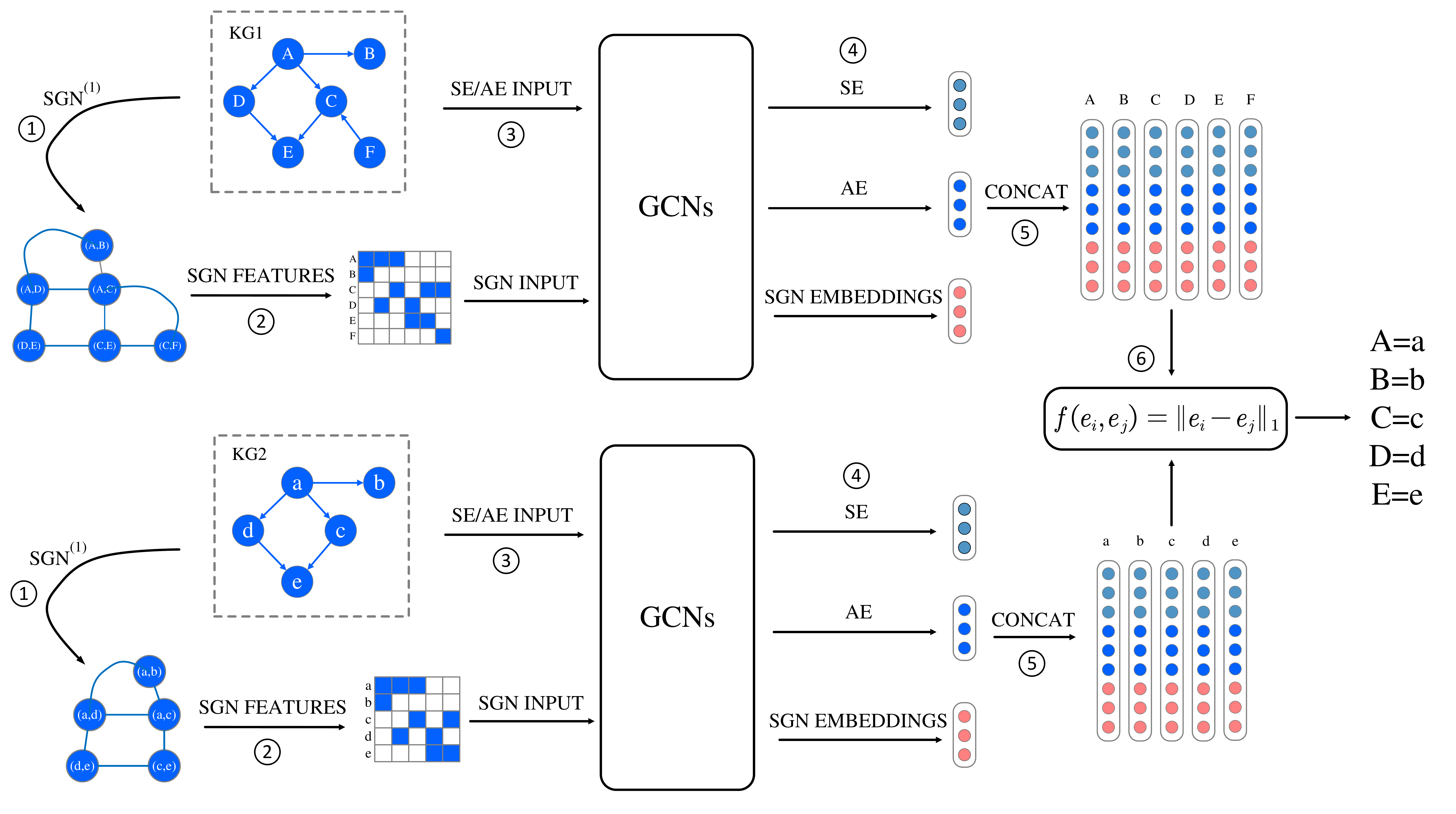}
\caption{The framework of sub-GCN, which consists of six steps. First, convert the original KG into a first-order SGN; secondly, extract first-order subgraph features for entities of the original network; third, obtain the structure and attribute features of the original knowledge graph as input; fourth, obtain the structure, attributes and SGN embedding vectors by training GCN models; fifth, concatenate the three embedding vectors according to different weights; sixth, align the entities in the two knowledge graphs according to the scoring function.} \label{fig1}
\end{figure}

\section{Methodology}
\label{sec:Methodology}
The graph alignment framework proposed in this paper is shown in Figure 1, which is divided into three parts: subgraph feature extraction, entity embedding model and entity alignment model. For a given $KG_1$ and $KG_2$ knowledge graph in two different languages and a set of known alignment entity pairs $S=\{(e_{i1},e_{i2})\}^m_{i=1}$. The subgraph feature extraction part uses the first-order SGN algorithm to extract the subgraph network from the original KG, and then performs the subgraph feature encoding for each entity. Entity embedding uses the GCN model to encode the entities in the KG, and embeds entities from different languages into a unified vector space. After training, the distance between equivalent entities will be as close as possible. Finally, the candidate entities are ranked by a predefined distance function to find the corresponding equivalent entity of each entity.

\subsection{Problem Definition}
The knowledge graph stores knowledge in the real world in the form of triples. In this article, we divide the graph into two types: relation triples and attribute triples. For example, in the DBP15K data set, $(Ren'Py, operatingSystem, Linux)$ is a relational triplet, and its format is $(entity_1, relation, entity_2)$. $(GaryLocke, \\dateOfBirth, 1950)$ is an attribute triplet, The format is $(entity, attribute, value)$. Formally, we express the knowledge graph as $KG={E, R, A, T_R, T_A}$, where E, R, A represent the set of entities, relations and attributes, and $T_R \subseteq (E \times R \times E)$ represents the relationship triplet set, $T_A \subseteq (E \times A \times V)$  represents the relationship triplet set, and V represents the set of attribute values.

For $KG_{1}=\{E_1,R_1,A_1,T_1^R,T_1^A\}$ and $KG_{2}=\{E_2,R_2,A_2,T_2^R,T_2^A\}$ two knowledge graphs in different languages, we define the task of cross-language knowledge graph alignment as finding new alignment entity pairs in KG through existing known entities. This experiment uses a set of pre-aligned entity pairs $S=\{(e_{i1},e_{i2}) | e_{i1} \in E_1,e_{i2} \in E_2\}_{(i=1)}^m$ to train the GCN model, and then discover new aligned entity pairs based on the distance function , where the pre-aligned entity pairs are constructed by cross-language links in DBpedia.

\subsection{Enhanced Structure Embedding Based On Subgraph Feature}
Subgraph Networks~\cite{xuan2019subgraph}  can effectively expand the structural feature space of the original network, and integrating this structural information can improve the performance of subsequent algorithms based on network structure. In the task of aligning cross-language knowledge graphs, the structures of the two graphs are very similar, and the equivalent entities often have the same subgraph structure characteristics.Therefore, extracting subgraph features to supplement the structural information of the original KG can effectively improve the accuracy of the entity alignment task. This paper extracts the first-order subgraph network of the original KG, and encodes the first-order subgraph feature for each entity. Specific steps are as follows:

\begin{itemize}
\item [$\bullet$]Detect subgraphs from the original network. In the method, the most basic sub-graphs (i.e. lines) are selected as sub-graphs because they are simple and relatively frequent in most networks.
\item [$\bullet$]Use subgraphs to construct SGN. After extracting enough subgraphs from the original network, a connection is established between them according to certain rules to construct SGN. Here, for simplicity, as long as two subgraphs share the same node or link in the original network, connect them.
\item [$\bullet$]Use SGN to construct the subgraph features of the original network. The new node in SGN is regarded as the encoding object of ont-hot encoding. If the entity in the original knowledge graph belongs to one of the subgraphs, it is marked as 1 at the corresponding position.
\end{itemize}

\begin{algorithm}
  \caption{ Constructing Subgraph Feature.}  
  \LinesNumbered
   \KwIn {A Knowledge Graph $KG(E, R, T_R)$ with entity set $E$, relation set $R$ and relation triples $T_R \subseteq (E \times R \times E)$.}
   \KwOut {Subgraph Feature Matrix $h_{sgn}$.}
    Initialize a network $G(V^{\prime},E^{\prime})$ with node set $V^{\prime} \subseteq   E$ and link set $E^{\prime} \subseteq  (V^{\prime} \times V^{\prime})$ by $KG(E, R, T_R)$\;
    $L=\{l_1,l_2,...,l_3\}$ $\gets$ extracting lines as subgraphs\;
          \For{$ i,j \in L$}
      {
          \If {$i,j$ share a common node  $\in V^{\prime}$}
         {
          add the link$(i,j)$ into $\widetilde{E}$
         }
      }
    get the $SGN^{(1)}$ denoted by $G(L,\widetilde{E})$\;
    \For{$ each\;e \in E$}
    {
    \If {if any subgraph $\in L$ contains $e$}
    {
        marked subgraph feature vector as 1 at the corresponding position
       } 
      }
    \Return Subgraph Feature Matrix $h_{sgn}$;  
\end{algorithm} 

\subsection{GCN-based Entity Embedding}
In this paper, three GCN models was trained by the structural features, attribute features and subgraph features of the knowledge graph. The parameters of the three GCN models are shown in Table 1. Each GCN model consists of multiple fixed GCN layers, and the $l+1$ layer depends on the output of the previous layer, with the specific 
convolution calculated as follows:
\begin{equation}
\begin{smallmatrix}
[H_s^{(l+1)};H_a^{(l+1)};H_{sgn}^{(l+1)}]=\sigma(\widehat{Q}^{-\frac{1}{2}}\widehat{P}\widehat{Q}^{-\frac{1}{2}}[H_s(l)W_s(l);H_a(l)&W_a(l);H_{sgn}(l)W_{sgn}(l)]),
\end{smallmatrix}
\end{equation}


Where $P$ in this paper adopts the adjacency matrix calculation method for knowledge graph proposed by Wang. $H_s(l)$, $H_a(l)$ and $H_{sgn}(l)$ are the vertex feature matrix, attribute feature matrix and subgraph feature matrix input To the first layer of the GCN model, $W_s(l)$, $W_a(l)$ and $W_{sgn}(l)$ are the weight matrices of the neural network of the first layer of various features.
The GCN model embeds the structural features and attribute features of different language entities into a unified vector space. In order to improve accuracy, this paper designs the following embedding method to enhance structural features, which reduces the difference between equivalent entities.

\subsection{Model Training}
In order to make the embedding vectors of equivalent entities as close as possible in the vector space, this paper uses a set of pre-aligned entity pairs $S$ as the training set of the GCN model, and trains model by minimizing the following loss function based on marginal ranking: 

\begin{equation}
\begin{smallmatrix}
L_s=\sum_{(e,v)\in S}\sum_{(e^{'},v^{'})\in S}[f(h_s(e),h_s(v))+\gamma_s-f(h_s(e^{'}),h_s(v^{'}))]_{+},
\end{smallmatrix}
\end{equation}
\begin{equation}
\begin{smallmatrix}
L_a=\sum_{(e,v)\in S}\sum_{(e^{'},v^{'})\in S}[f(h_a(e),h_a(v))+\gamma_a-f(h_a(e^{'}),h_a(v^{'}))]_{+},
\end{smallmatrix}
\end{equation}
%
\begin{equation}
\begin{smallmatrix}
L_{sgn}=\sum_{(e,v)\in S}\sum_{(e^{'},v^{'})\in S}[f(h_{sgn}(e),h_{sgn}(v))+\gamma_{sgn}-f(h_{sgn}(e^{'}),h_{sgn}(v^{'}))]_{+},
\end{smallmatrix}
\end{equation}
where $[x]_{+}=max\{0,x\}$, $f(x,y)=\left\|{x-y}\right\|_1$, $S_{(e,v)}^{'}$ is to randomly replace the aligned entity pairs. The negative sample set constructed by an entity in $(e,v)$. The replaced entity is randomly selected from $KG_1$ and $KG_2$. $\gamma_s$, $\gamma_a$, $\gamma_{sgn}>0$ is a hyperparameter used to control the degree of positive and negative alignment entities. $L_s$, $L_a$, and $L_{sgn}$ are the loss functions of structural embedding, attribute embedding, and subgraph embedding, respectively. They are independent of each other, so they will be optimized using stochastic gradient descent respectively.

\subsection{Knowledge Graph Entity Alignment Prediction}
Knowledge Graph entity alignment is predicted based on the distance between two entities in vector space. For the entity $e_i$ in $KG_1$ and the entity $e_j$ in $KG_2$, the distance between them is calculated by the following formula:

\begin{equation}
\begin{smallmatrix}
D(e_i,e_j)=&\alpha\frac{f(h_s(e_i),h_s(e_j))}{d_s}+\beta\frac{f(h_a(e_i),h_a(e_j))}{d_a}+\gamma\frac{f(h_{sgn}(e_i),h_{sgn}(e_j))}{d_{sgn}},
\end{smallmatrix}
\end{equation}

where $h_s(e)$, $h_a(e)$, $h_{sgn}(e)$ represent vectors after structural embedding, attribute embedding, and first-order subgraph embedding, $d$ represents their dimensions. $\alpha$, $\beta$, $\gamma$ are the parameter to balance the three embedding vectors and need to satisfy $\alpha+\beta+\gamma=1$. For entity $e_i$ in $KG_1$, this method calculates the distance from all entities in $KG_2$ to $e_i$ and ranks them as candidate equivalent entities. In addition, the alignment from $KG_2$ to $KG_1$ can also be performed. The results of the alignment in both directions are given in the next section.

\begin{table}
\vspace{-2mm}
\centering
\caption{Parameters of Three GCNs}\label{tab1}
\resizebox{\textwidth}{!}{
\begin{tabular}{|l|l|l|l|}
\hline
\multicolumn{2}{|c|}{Parameter}                                                               & \multicolumn{1}{c|}{$KG_1$}       & \multicolumn{1}{c|}{$KG_2$}      \\ \hline
\multicolumn{1}{|c|}{\multirow{4}{*}{$GCN_{SE}$}} & Initial Structure Feature Matrices              & $H_{s1}^{(0)} \in \mathbb{R}^{\vert E_1\vert \times d_s}$           & $H_{s2}^{(0)} \in \mathbb{R}^{\vert E_2\vert \times d_s}$          \\ \cline{2-4} 
\multicolumn{1}{|c|}{}                      & Weight Matrix for Structure Features in Layer 1 & \multicolumn{2}{c|}{$W_s^{(1)} \in \mathbb{R}^{d_s \times d_s}$} \\ \cline{2-4} 
\multicolumn{1}{|c|}{}                      & Weight Matrix for Structure Features in Layer 2 & \multicolumn{2}{c|}{$W_s^{(2)} \in \mathbb{R}^{d_s \times d_s}$} \\ \cline{2-4} 
\multicolumn{1}{|c|}{}                      & Output Structure Embeddings                     &$H_{s1}^{(2)} \in \mathbb{R}^{\vert E_1\vert \times d_s}$           &$H_{s2}^{(2)} \in \mathbb{R}^{\vert E_2\vert \times d_s}$           \\ \hline
\multirow{4}{*}{$GCN_{AE}$}                       & Initial Attribute Feature Matrices              & $H_{a1}^{(0)} \in \mathbb{R}^{\vert E_1\vert \times \vert A_1\vert}$         &$H_{a2}^{(0)} \in \mathbb{R}^{\vert E_2\vert \times \vert A_2\vert}$           \\ \cline{2-4} 
                                            & Weight Matrix for Attribute Features in Layer 1 & \multicolumn{2}{c|}{$W_a^{(1)} \in \mathbb{R}^{\vert A_1\vert \times d_a}$} \\ \cline{2-4} 
                                            & Weight Matrix for Attribute Features in Layer 2 & \multicolumn{2}{c|}{$W_a^{(2)} \in \mathbb{R}^{d_a \times d_a}$} \\ \cline{2-4} 
                                            & Output Attribute Embeddings                     &$H_{a1}^{(2)} \in \mathbb{R}^{\vert E_1\vert \times d_a}$           &$H_{a1}^{(2)} \in \mathbb{R}^{\vert E_2\vert \times d_a}$           \\ \hline
\multirow{4}{*}{$GCN_{SGN}$}                       & Initial Subgraph Feature Matrices               &$H_{sgn1}^{(0)} \in \mathbb{R}^{\vert E_1\vert \times \vert SGN\vert}$           &$H_{sgn2}^{(0)} \in \mathbb{R}^{\vert E_2\vert \times \vert SGN\vert}$           \\ \cline{2-4} 
                                            & Weight Matrix for Subgraph Features in Layer 1  & \multicolumn{2}{c|}{$W_{sgn}^{(1)} \in \mathbb{R}^{\vert SGN\vert \times d_{sgn}}$} \\ \cline{2-4} 
                                            & Weight Matrix for Subgraph Features in Layer 2  & \multicolumn{2}{c|}{$W_{sgn}^{(2)} \in \mathbb{R}^{d_{sgn} \times d_{sgn}}$} \\ \cline{2-4} 
                                            & Output Subgraph Embeddings                      & $H_{sgn1}^{(2)} \in \mathbb{R}^{\vert E_1\vert \times d_{sgn}}$          &$H_{sgn2}^{(2)} \in \mathbb{R}^{\vert E_2\vert \times d_{sgn}}$           \\ \hline
\end{tabular}}
 \vspace{-4mm}
\end{table}

\section{Experiments}
\label{sec: Experiments}
\subsection{Experimental Setting}
Here, we will explain the experimental setting. To prove the effectiveness of Sub-GCN, we performed an entity alignment task in the DBP15K dataset and compared it with some of the latest baseline methods. For all alignment models, we use 30\% of the pre-aligned entity links as the training set, and the remaining 70\% for testing. The parameters in the sub-GCN are set to $d_s=200$, $d_{sgn}=d_a=100$; $\gamma_s=\gamma_{sgn}=\gamma_a=3$; $epochs=5000$; $\alpha=0.72$; $\beta=0.2$;$\gamma=0.08$;the number of negative samples for each positive seed parameter $k=20$. The results GCN[Wang] obtained from Wang et al.~\cite{wang2018cross}, and *  marked result parameter settings are as follows: $d_s$=200, $d_a$=100; $\gamma_s=\gamma_a=3$; $epochs=5000$; $\alpha=0.8$; $\beta=0.2$; $k=20$.

\subsection{DataSet}
The data set in the experiment is DBP15K, which was constructed by Sun~\cite{sun2017cross} and others. The DBP15K data set comes from DBpedia, which is a large-scale multilingual knowledge map from which DBP15K extracts a subset of the KG in Chinese, English, Japanese, and French. The description of the data set is shown in Table 2. Each data set contains two knowledge graphs in different languages and 15,000 equivalent entity links.

\begin{table}
\centering
\caption{Details of the datasets}\label{tab2}
\resizebox{\textwidth}{!}{
\begin{tabular}{lllllll}
\hline\hline
\multicolumn{2}{l}{Datasets}              & Entities & Rleations & Attributes & Rel. triple & Attr. triples \\
\hline
\multirow{2}{*}{DBP15K\_ZH-EN} & Chinese  & 66469    & 2830      & 8113       & 153929      & 379684        \\
                               & English  & 98125    & 2317      & 7173       & 237674      & 567755        \\
\hline
\multirow{2}{*}{DBP15K\_JA-EN} & Japanese & 65744    & 2043      & 5882       & 164373      & 354619        \\
                               & English  & 95680    & 2096      & 6066       & 233319      & 497230        \\
\hline
\multirow{2}{*}{DBP15K\_FR-EN} & French   & 66858    & 1379      & 4547       & 192191      & 528665        \\
                               & English  & 105889   & 2209      & 6422       & 278590      & 576543\\
\hline\hline
\end{tabular}}
 \vspace{-6mm}
\end{table}

\subsection{Result}
In the experiment, we mainly compared our method with the recent work based on GCN ~\cite{wang2018cross}, and also compared some methods based on TransE embedding, such as JE, MTransE and JAPE models. The experimental results are shown in Table 3,4,5, where SE means that only structural information is used for embedding, SE+AE means that both structure and attribute information are used for embedding. The sub-GCN is the model proposed in this paper which uses three types of information: structure, attributes and subgraph features. We use Hits@k as an evaluation indicator to evaluate the performance of all methods. Hits@k represents the hit rate of the correct entity among the top $k$ candidate entities. 
\begin{table}
\centering
\caption{Results comparison of cross-lingual KG alignment (ZH-EN)}\label{tab3}
\resizebox{\textwidth}{!}{
\begin{tabular}{clllllll}
\hline\hline
\multicolumn{2}{c}{\multirow{2}{*}{DBP15K\_ZH-EN}} & \multicolumn{3}{c}{ZH-EN}  & \multicolumn{3}{c}{EN-ZH}  \\
\multicolumn{2}{c}{}                               & Hits@1 & Hits@10 & Hits@50 & Hits@1 & Hits@10 & Hits@50 \\
\hline
\multicolumn{2}{c}{JE}                             & 21.27  & 42.77   & 56.74   & 19.52  & 39.36   & 53.25   \\
\hline
\multicolumn{2}{c}{MTransE}                        & 30.83  & 61.41   & 79.12   & 24.78  & 52.42   & 70.45   \\
\hline
\multirow{2}{*}{JAPE}              & SE w/o neg.           & 38.34  & 68.86   & 84.07   & 31.66  & 59.37   & 76.33   \\
                                   & SE            & 39.78  & 72.35   & 87.12   & 32.29  & 62.79   & 80.55   \\
                                   & SE+AE         & 41.18  & 74.46   & \textbf{88.90}   & 40.15  & 71.05   & \textbf{86.18}   \\
                                   \hline
\multirow{2}{*}{GCN{[}Wang{]}}     & SE            & 38.42  & 70.34   & 81.24   & 34.43  & 65.68   & 77.03   \\
                                   & SE+AE         & 41.25  & 74.38   & 86.23   & 36.49  & 69.94   & 82.45   \\
                                   \hline
\multirow{2}{*}{*GCN{[}Wang{]}}     & *SE            & 40.22  & 69.53   & 78.76   & 37.81  & 67.30   & 76.95   \\
                                   & *SE+AE         & 44.71  & 75.89   & 86.51   & 42.14  & 73.92   & 84.78   \\
                                   \hline
\multicolumn{2}{c}{sub-GCN}     & \textbf{45.01}  & \textbf{76.48}   & 87.02   & \textbf{42.97}  & \textbf{75.30}   & 85.72  \\
\hline\hline
\end{tabular}}
\vspace{-7mm}
\end{table}

\begin{table}
\centering
\caption{Results comparison of cross-lingual KG alignment (JA-EN)}\label{tab4}
\resizebox{\textwidth}{!}{
\begin{tabular}{clllllll}
\hline\hline
\multicolumn{2}{c}{\multirow{2}{*}{DBP15K\_ZH-EN}} & \multicolumn{3}{c}{ZH-EN}  & \multicolumn{3}{c}{EN-ZH}  \\
\multicolumn{2}{c}{}                               & Hits@1 & Hits@10 & Hits@50 & Hits@1 & Hits@10 & Hits@50 \\
\hline
\multicolumn{2}{c}{JE}                             & 18.92  & 39.97   & 54.24   & 17.80  & 38.44   & 52.48   \\
\hline
\multicolumn{2}{c}{MTransE}                        & 27.86  & 57.45   & 75.94   & 23.72  & 49.92   & 67.93   \\
\hline
\multirow{2}{*}{JAPE}              & SE w/o neg.           & 33.10  & 63.90   & 80.80   & 29.71  & 56.28   & 73.84   \\
                                   & SE            & 34.27  & 66.39   & 83.61   & 31.40  & 60.80   & 78.51   \\
                                   & SE+AE         & 36.25  & 68.50   & 85.35   & 38.37  & 67.27   & 82.65   \\
                                   \hline
\multirow{2}{*}{GCN{[}Wang{]}}     & SE            & 38.21  & 72.49   & 82.69   & 36.90  & 68.50   & 79.51   \\
                                   & SE+AE         & 39.91  & 74.46   & 86.10   & 38.42  & 71.81   & 83.72   \\
                                   \hline
\multirow{2}{*}{*GCN{[}Wang{]}}    & *SE            & 41.90  & 72.73   & 81.43   & 39.72  & 69.06   & 77.82   \\
                                   & *SE+AE         & 45.19  & 77.10   & 87.63   & 42.83  & 74.02   & 85.96   \\
                                   \hline
\multicolumn{2}{c}{sub-GCN}       & \textbf{45.46}  & \textbf{78.15}   & \textbf{88.98}   & \textbf{43.50}  & \textbf{75.46}   & \textbf{87.18}  \\
\hline\hline
\end{tabular}}
\vspace{-7mm}
\end{table}

\begin{table}
\centering
\caption{Results comparison of cross-lingual KG alignment (FR-EN)}\label{tab5}
\resizebox{\textwidth}{!}{
\begin{tabular}{clllllll}
\hline\hline
\multicolumn{2}{c}{\multirow{2}{*}{DBP15K\_FR-EN}} & \multicolumn{3}{c}{FR-EN}  & \multicolumn{3}{c}{EN-FR}  \\
\multicolumn{2}{c}{}                               & Hits@1 & Hits@10 & Hits@50 & Hits@1 & Hits@10 & Hits@50 \\
\hline
\multicolumn{2}{c}{JE}                             & 15.38  & 38.84   & 56.50   & 14.61  & 47.25   & 54.01   \\
\hline
\multicolumn{2}{c}{MTransE}                        & 24.41  & 55.55   & 74.41   & 21.26  & 50.60   & 69.93   \\
\hline
\multirow{2}{*}{JAPE}              & SE w/o neg.   & 29.55  & 62.18   & 79.36   & 25.40  & 56.55   & 74.96   \\                                   & SE            & 29.63  & 64.55   & 81.90   & 26.55  & 69.39   & 78.71   \\
                                   & SE+AE         & 32.39  & 66.68   & 83.19   & 32.97  & 65.91   & 82.38   \\
                                   \hline
\multirow{2}{*}{GCN{[}Wang{]}}     & SE            & 36.51  & 73.42   & 85.93   & 36.08  & 72.37   & 85.44   \\
                                   & SE+AE         & 37.29  & 74.49   & 86.73   & 42.50  & 76.78   & 87.92   \\
                                   \hline
\multirow{2}{*}{*GCN{[}Wang{]}}     & *SE            & 42.07  & 76.34   & 85.66   & 40.33  & 73.71   & 83.90   \\
                                   & *SE+AE         & 43.44  & 79.21   & 88.93   & 42.44  & 76.85   & 87.88   \\
                                   \hline
\multicolumn{2}{c}{sub-GCN}        & \textbf{43.67}  & \textbf{79.65}   & \textbf{89.68}   & \textbf{42.93}  & \textbf{77.38}   & \textbf{88.93}  \\
\hline\hline
\end{tabular}}
\end{table}

\subsubsection{\textbf{sub-GCN vs. JAPE}}
According to the data in the table, the sub-GCN alignment model is superior to JAPE in the alignment of the $DBP15K_{JA-EN}$ and $DBP15K_{FR-EN}$. This shows that in the cross-language knowledge graph, the GCN model based on the graph structure is easier to capture the features of the entity than the model based on the TransE embedding model.
The effect of the sub-GCN alignment model on the alignment of $DBP15K_{ZH-EN}$ is not better than that of JAPE on the hits@50 index. This may be because the embedding of the JAPE model can use the triples of relations and attributes at the same time, so better results can be obtained on data sets with multiple relations and attributes. But this also increases the cost of pre-labeling, and the training of the model requires pre-aligning the entire triplet. In this regard, GCN-based embedding only needs to label aligned entities, which is suitable for the alignment of large knowledge graphs.

\begin{figure}
\centering
\includegraphics[width=\textwidth]{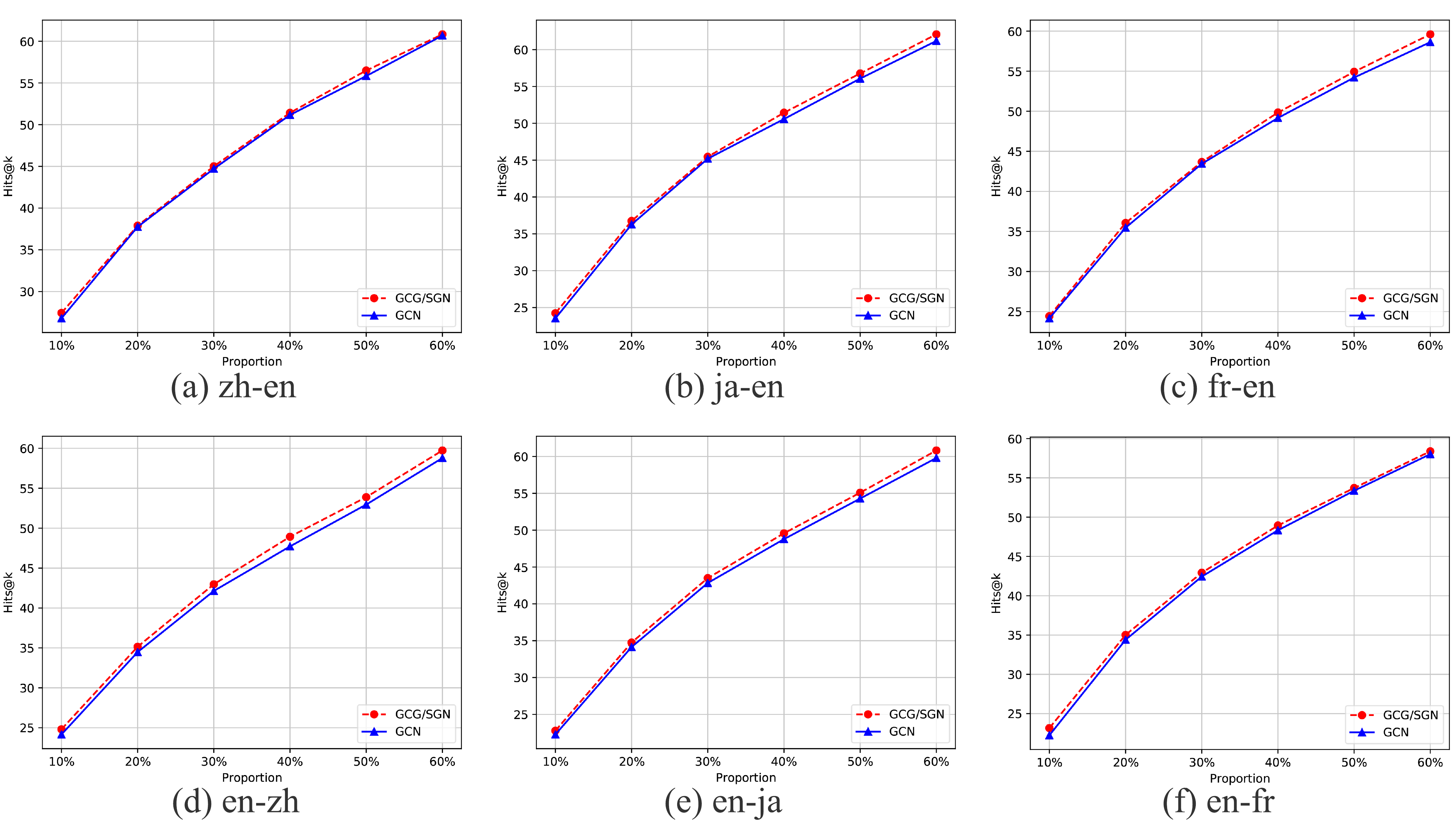}
\caption{ GCN(AE+SE) and sub-GCN using different sizes of training data;vertical coordinates: Hits@1} \label{fig2}
\end{figure}

\begin{figure}
\centering
\includegraphics[width=\textwidth]{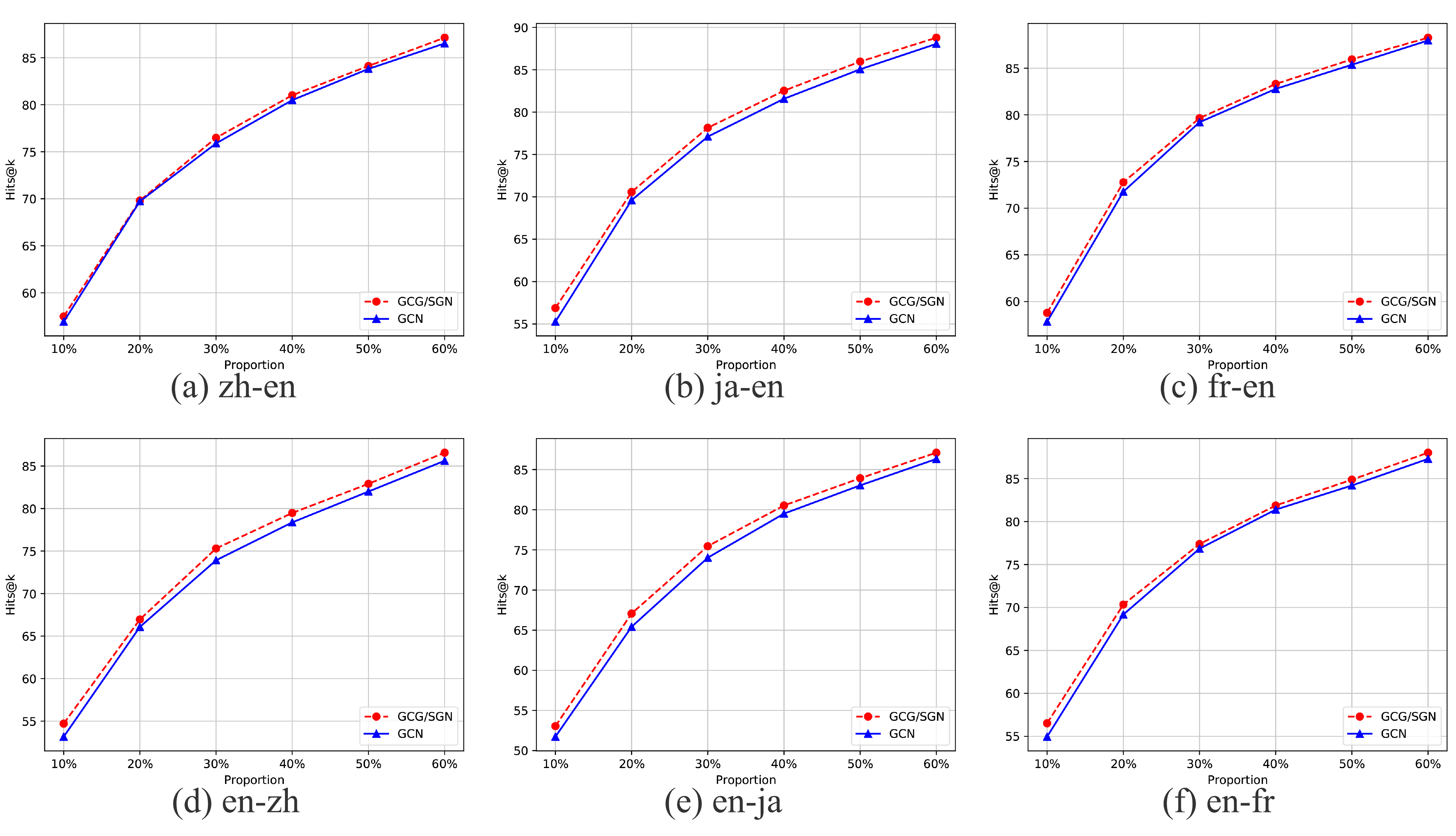}
\caption{ GCN(AE+SE) and sub-GCN using different sizes of training data;vertical coordinates: Hits@10} \label{fig3}
\end{figure}

\begin{figure}
\centering
\includegraphics[width=\textwidth]{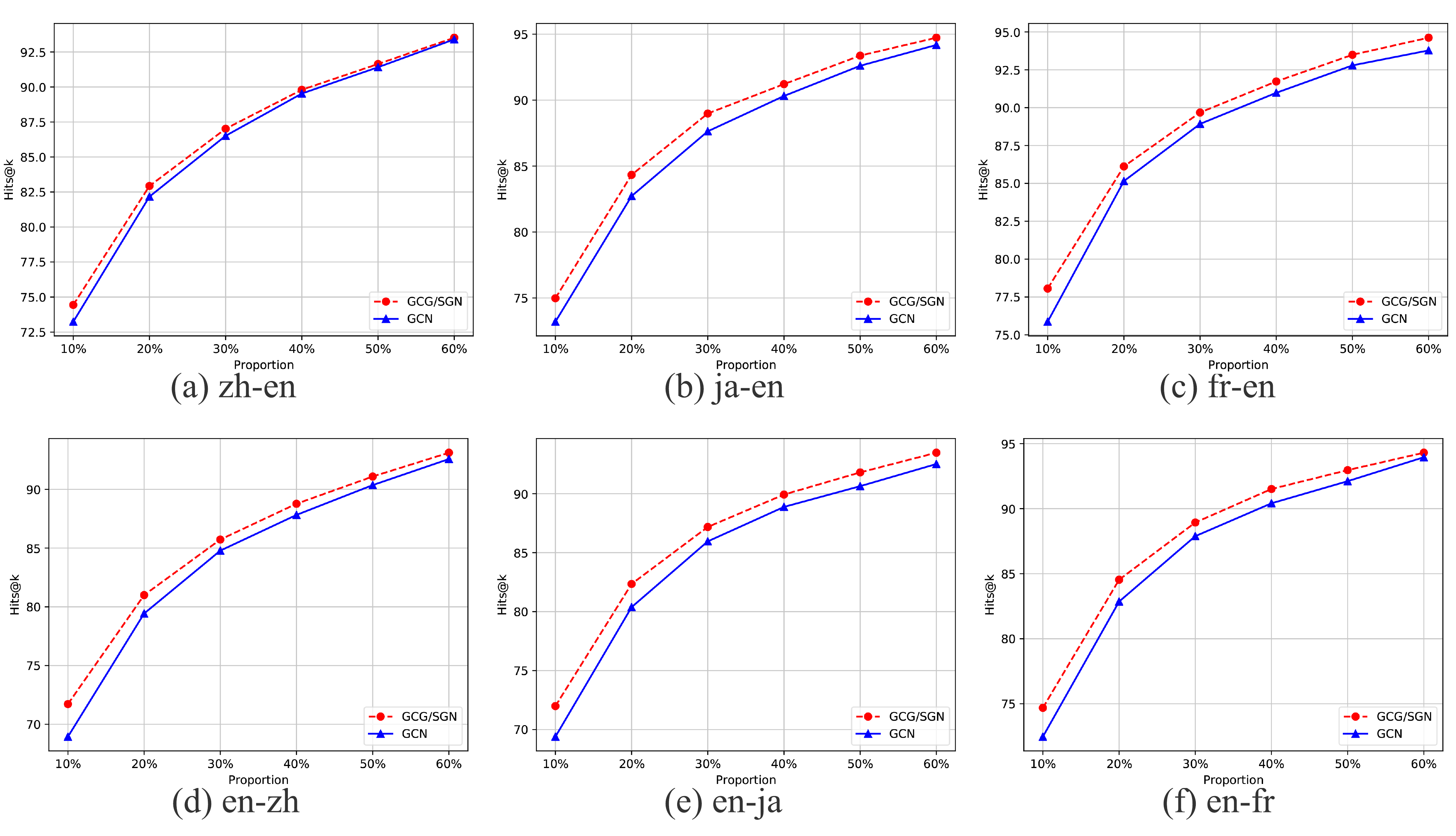}
\caption{ GCN(AE+SE) and sub-GCN using different sizes of training data;vertical coordinates: Hits@50} \label{fig4}
\end{figure}

\subsubsection{\textbf{sub-GCN vs. GCN(SE+AE)}}
Compared with GCN (SE+AE), sub-GCN outperforms GCN (SE+AE) on the three datasets. After extracting the first-order subgraph of the knowledge graph to enhance the structural information, the alignment ability of the GCN model has been improved. Experiments show that in the task of aligning cross-language knowledge graphs, the subgraph feature can effectively expand and expand the structural information of the original graph to obtain a better alignment effect.

\subsubsection{\textbf{GCN vs. sub-GCN using different sizes of training data}}
To study the effect of different scale training sets on experimental results, we used different numbers of pre-aligned entity pairs as training sets for the GCN[Wang] and sub-GCN models. The experiment selected different proportions of pre-aligned entity pairs as the training set, which ranges from 10\% to 60\% and with a step length of 10\%. The remaining pre-aligned entity pairs are used as test data. The experimental results show that the effect of the GCN[Wang] and sub-GCN models is positively related to the ratio of pre-aligned entity pairs on the three indicators of $Hits@1$, $Hits@10$, $Hits@50$. Moreover, no matter how many proportional training sets are selected, the method of this paper is always better than that of GCN[Wang], especially in the $DBP15K_{JA-EN}$ data set.

\section{Conclusion}
\label{sec:Conclusion}
This paper designs and implements a cross-language knowledge graph entity alignment method based on subgraph features. The method consists of subgraph feature extraction, entity embedding and entity alignment prediction, and performed entity alignment tasks on multiple real multilingual knowledge graphs. The experimental results show that the method can enhance the structural information of the original KG and obtain a higher hit rate, which provides a new solution for the alignment of the  cross-language knowledge graph. In our future work, we will explore effective and fast subgraph feature extraction methods for second or higher order.

\subsubsection*{Acknowledgments}This work was partially supported by National Natural Science Foundation of China under Grant 62103374 and by Basic Public Welfare Research Project of Zhejiang Province under Grant LGF20F020016.

\bibliographystyle{splncs04}
\bibliography{GCN-Align}

\end{document}